\documentclass[12pt]{article}
\usepackage{graphicx}

\usepackage{url} 
\usepackage{amsmath}
\usepackage{amsthm}
\usepackage{amssymb}
\usepackage{tabularx}
\usepackage{multirow}
\usepackage{hyperref}
\newcommand{\mbf}[1]{\boldsymbol{#1}}
\newcommand{\mbb}[1]{\mathbb{#1}}
\newcommand{\mcal}[1]{\mathcal{#1}}
\newcommand{\mrm}[1]{\textrm{#1}}
\newcommand\numberthis{\addtocounter{equation}{1}\tag{\theequation}}

\newcommand{\norm}[1]{\left\|{#1}\right\|}
\newcommand{\abs}[1]{\left|{#1}\right|}

\newtheorem{thm}{Theorem}
\newtheorem{mydef}{Definition}
\newtheorem{lemma}{Lemma}
\newtheorem{prop}{Proposition}

\newcommand{\blind}{0}

\begin{document}

\def\spacingset#1{\renewcommand{\baselinestretch}%
{#1}\small\normalsize} \spacingset{1}


\if0\blind
{
  \title{\bf Consistency of Neural Networks with Regularization}
  \author{Xiaoxi Shen\\
    Department of Math, Texas State university\\
    and \\
    Jinghang Lin\thanks{Two authors are equally contributed to this paper. If you have any questions, please email to jinghang.lin@yale.edu}\hspace{.2cm} \\
    Department of Biostatistics, Yale University}
  \maketitle
} \fi

\if1\blind
{
  \bigskip
  \bigskip
  \bigskip
  \begin{center}
    {\LARGE\bf Title}
\end{center}
  \medskip
} \fi

\bigskip
\begin{abstract}
Neural networks have attracted a lot of attention due to its success in applications such as natural language processing and computer vision. For large scale data, due to the tremendous number of parameters in neural networks, overfitting is an issue in training neural networks. To avoid overfitting, one common approach is to penalize the parameters especially the weights in neural networks. Although neural networks has demonstrated its advantages in many applications, the theoretical foundation of penalized neural networks has not been well-established. Our goal of this paper is to  propose the general framework of neural networks with regularization and prove its consistency. Under certain conditions, the estimated neural network will converge to true underlying function as the sample size increases. The method of sieves and the theory on minimal neural networks are used to overcome the issue of unidentifiability for the parameters. Two types of activation functions: hyperbolic tangent function(Tanh) and rectified linear unit(ReLU) have been taken into consideration. Simulations have been conducted to verify the validation of theorem of consistency.
\end{abstract}

\noindent%
\vfill

\newpage
\spacingset{1.8} 
\section{Introduction}
Neural network-based methods have achieved tremendous success in many areas, such as computer vision and natural language processing. In those applications, neural networks have a tendency to be more complicated. Some popular neural networks methods(e.g. Alexnet, VGG, ResNet) have millions of parameters which may have issue of overfitting \cite{He,Krizhevsky2012ImageNetNetworks,Simonyan2015VERYRECOGNITION}. Many regularization approaches are proposed to address it, such as dropout and early-stopping. For more information, readers could refer to the deep learning book written by Ian Goodfellow, Yoshua Bengio and Aaron Courville \cite{IanGoodfellowandYoshuaBengioandAaronCourville2016DeepLearning}. In statistics, penalization is a common approach to reduce the complexity of the model and hence alleviate the effect of overfitting. Many researchers focused on the application of neural networks. Although it has been
shown that penalized neural networks have considerable accuracy in supervised learning, the lack of enough statistical properties has left penalized neural networks, like many other neural
networks, as "black boxes". Since statistical properties of penalized neural networks are not well studied, it is worthwhile to explore whether penalized neural networks possess nice asymptotic properties, such as consistency.

One of the advantages for neural networks is that they are universal approximators. Between 1990 and 2000, many theories on the universal approximation theorem for neural networks have been established using various methods such as Stone-Weierstrass Theorem or Hahn-Banach Theorem \cite{ cybenko1989approximation, funahashi1989approximate, Hornik1989MultilayerApproximators,leshno1993multilayer}. Later on, approximation rates for neural networks were established \cite{barron1993universal, hornik1994degree, makovoz1996random, mhaskar1996neural}. Pinkus (1999) \cite{pinkus1999approximation} provides a comprehensive summary on the results of universal apprximations theorem and approximation rate for shallow neural networks. 

Besides the mathematical results on approximation accuracy, statistical properties based on shallow 
neural networks have also been established. Barron (1994) \cite{barron1994approximation} showed 
that the $L_2$-distance between a shallow neural network estimator with $n$ hidden units and the 
underlying function is of the order $\mcal{O}(n^{-1/2})$. This is an important finding as the rate does 
not depend on the dimension of the input features. The curse of dimensionality 
\cite{stone1980optimal} is a well-known issue for nonparametric estimators. The results in 
\cite{barron1994approximation} indicate that neural networks estimators can avoid this issue. Chen 
and White (1999) \cite{chen1999improved} obtained an even faster rate than the one obtained in 
\cite{barron1994approximation} but the rate is depends on the input dimension and such a rate 
converges to Barron's rate as the input dimension tends to infinity. Recently, Shen et al. (2019) 
\cite{shen2019asymptotic} provides a comprehensive study on the statistical properties, including the 
consistency, rate of convergence and asymptotic normality, for shallow neural networks.

The aforementioned work mainly focused on the sigmoid activation function. However, due to the problem of vanishing gradient, sigmoid activation function is not a popular choice for deep neural networks. The rectified linear unit (ReLU) is the activation function that is widely used in deep learning applications. In recent years, many studies have been completed to extend the theories for shallow neural networks to their deep counterparts. For example, a series work by Yarotsky \cite{yarotsky2017error,yarotsky2018optimal,yarotsky2020phase} have established the approximation rates for deep neural networks with ReLU activation function for functions in the unit ball of a Sobolev space. In \cite{yarotsky2020phase}, he pointed out the phase transition in terms of the approximability of shallow neural networks and deep neural networks. Fabozzi et al. (2019) \cite{Fabozzi2019Towards} extended the theories in \cite{shen2019asymptotic} to deep neural networks with ReLU activation function. Asymptotic properties of deep neural 
networks in semiparametric setting are obtained\cite{farrell2021deep}.  
Fully connected deep neural network regression estimates in nonparametric setting has been  analyzed\cite{kohler2021rate,schmidt2020nonparametric}.

The goal of this paper is to fill in the gap of theoretical work in neural networks with regularization.
Since neural networks are nonlinear function under nonparametric regression setting, it can be viewed as a form of sieve estimator. We derive the consistency of penalized neural networks in
a general framework. By applying a basic inequality regarding about objective function, we 
derive a upper bound for least square loss function. Penalized neural networks requires more 
care than its machine learning counterparts, due to severe nonlinearity and heavy parametrization.  Two types of activation functions: hyperbolic tangent function(Tanh) and Rectified Linear Unit(ReLu) are considered. We use the method of sieves to narrow down the parameter space. For different type of activation function, the penalty term

The paper is organized as follows. Section \ref{sec: consistency, general result} gives a  general result of consistency for any function with regularization. Section  presents the parameterization of penalized neural networks with two activation settings: Tanh 
and ReLu. Section 5 explores the
validity of the theoretical results by conducting simulations. Section 6 concludes our paper.

\section{Consistency on Nonparametric Penalized Least Square}\label{sec: consistency, general result}
We consider the general nonparametric regression problem:
\begin{align}
    Y_i = f_{0}(X_i) + \epsilon_i
\end{align}
$X_1,..,X_n$ are i.i.d from distribution $P$ and $X \in \mathcal{X} \subset \mathbb{R}^d$, where $\mathcal{X}$ is a compact set in $\mathbb{R}^{d}$. $\epsilon_1,.., \epsilon_n$ are i.i.d random error with $E[\epsilon] = 0$ and $||\epsilon||_{p,1}=\int_0^\infty(\mbb{P}(\abs{\epsilon}>t))^{1/p}dt < \infty$ for some $p\leq 2$. $f_0 \in \mathcal{F}$ is the underlying function to be estimated.
To get the estimator, we minimize the following objective function.
\begin{align}\label{Eq: PLS}
    \tilde{\mathbb{Q}}_n(f)= \frac{1}{n}\sum_{i=1}^{n}(f(X_i)-Y_i)^{2}+\lambda_{n} J_{n}(f),
\end{align}
where $J_n(\cdot): \mathcal{F} \to \mathbb{R}^{+}$ is some penalty function and $\lambda_n$ is the regularization parameter. The approximate penalized sieve extremum estimator is considered:
\begin{align*}
    \Tilde{\mathbb{Q}}_n(\hat{f}_n) \leq \inf_{f\in \mathcal{F}_n} \Tilde{\mathbb{Q}}_n(f) + O(\eta_n), 
\end{align*}
where $\eta_n\to0$ as $n\to\infty$ and $\mathcal{F}_n \subset \mathcal{F}_{n+1}$ such that $\mathcal{F} = \bigcup_{n=1}^{\infty} \mathcal{F}_n$ in the sense that for any $f \in \mathcal{F}$, there exists $\pi_n f \in \mathcal{F}$ such that $\sup_{x \in \mathcal{X}} |\pi_n f(x) - f(x)| \to 0$ as $n \to \infty$. Here $\Tilde{Q}_n(f)$ is the penalized least square as defined in (\ref{Eq: PLS})

To begin with, we introduce a basic inequality for establishing the consistency of penalized nonparametric least square estimator.
\begin{lemma}[Basic Inequality]\label{Lm: Basic Ineq}
\begin{align*}
    \frac{1}{n} \sum_{i=1}^{n}(\hat{f}_n(X_i) - f_0(X_i))^2 &\leq \frac{1}{n} \sum_{i=1}^{n}(f_0(X_i) - \pi_{n}f_0(X_i))^2 + \frac{2}{n}\sum_{i=1}^{n}\epsilon_i(\hat{f}_n(X_i) - \pi_{n}f_0(X_i)) \\
    &+ \lambda_n [J_n(\pi_{n}f_0) - J_n(\hat{f}_n)] + O(\eta_n)
\end{align*}
\end{lemma}

\begin{proof}
By the definition of $\Tilde{\mathbb{Q}}_{n}(\hat{f}_n))$ and $Y_i = f_0(X_i) + \epsilon_i$, we have
\begin{align}
    \Tilde{\mathbb{Q}}_n(\hat{f}_n) &= \frac{1}{n}\sum_{i=1}^n(Y_i - \hat{f}(X_i))^2 + \lambda_n J_n(\hat{f}_n)\\
    & = \frac{1}{n}\sum_{i=1}^n(f_0(X_i) + \epsilon_i - \hat{f}(X_i))^2 + \lambda_n J_n(\hat{f}_n)\\
    & = \frac{1}{n}\sum_{i=1}^n\epsilon_{i}^{2} + \frac{2}{n}\sum_{i=1}^{n}\epsilon_i(f_0(X_i) - \hat{f}_n(X_i)) + \frac{1}{n} \sum_{i=1}^{n}(\hat{f}_n(X_i) - f_0(X_i))^2 + \lambda_n J_n(\hat{f}_n)
\end{align}
and
\begin{align}
    \Tilde{\mathbb{Q}}_n(\pi_{n}f_0) & = \frac{1}{n}\sum_{i=1}^n(Y_i - \pi_{n}f_{0}(X_i))^2 + \lambda_n J_n(\pi_{n}f_0) \\
    & = \frac{1}{n}\sum_{i=1}^n(f_0(X_i) + \epsilon_i - \pi_{n}f_{0}(X_i))^2 + \lambda_n J_n(\pi_{n}f_0) \\
    & = \frac{1}{n}\sum_{i=1}^n\epsilon_{i}^{2} + \frac{2}{n}\sum_{i=1}^{n}\epsilon_i(f_0(X_i) - \pi_{n}f_0(X_i))\\
    &+ \frac{1}{n} \sum_{i=1}^{n}(f_0(X_i) - \pi_{n}f_0(X_i))^2 + \lambda_n J_n(\pi_{n}f_0)
\end{align}
Since
\begin{align*}
    \Tilde{\mathbb{Q}}_n(\hat{f}_n) \leq & \inf_{f \in \mathcal{F}_n} \Tilde{\mathbb{Q}}_n(f) + O(\eta_n)\\
   & \leq \Tilde{\mathbb{Q}}_n(\pi_{n}f_0) + O(\eta_n),
\end{align*}
we have
\begin{align*}
    \frac{1}{n} \sum_{i=1}^{n}(\hat{f}_n(X_i) - f_0(X_i))^2 &\leq \frac{1}{n} \sum_{i=1}^{n}(f_0(X_i) - \pi_{n}f_0(X_i))^2 + \frac{2}{n}\sum_{i=1}^{n}\epsilon_i(\hat{f}_n(X_i) - \pi_{n}f_0(X_i)) \\
    &+ \lambda_n [J_n(\pi_{n}f_0) - J_n(\hat{f}_n)] + O(\eta_n)
\end{align*}
\end{proof}

Since $I :=\frac{1}{n} \sum_{i=1}^{n}(f_0(X_i) - \pi_{n}f_0(X_i))^2= ||\pi_{n}f_0 - f_0||^{2} \to 0$ by the denseness assumption on the sieve space and $\eta_n \to 0$ as $n \to \infty$, we only need to check the following two terms to get the consistency
\begin{itemize}
    \item $II :=\frac{2}{n}\sum_{i=1}^{n}\epsilon_i(\hat{f}_n(X_i) - \pi_{n}f_0(X_i))$
    \item $III := \lambda_n [J_n(\pi_{n}f_0) - J_n(\hat{f}_n)]$
\end{itemize}

\begin{lemma}[Convergence of Multiplier Process]\label{Lm: Convergence of Multiplier Process}
If $E[\sup_{f \in \mathcal{F}_n}|f(x)|] < \infty$, then
\begin{align*}
  &E[\sup_{f \in \mathcal{F}_n}|\frac{1}{\sqrt{n}}\sum_{i=1}^{n}\epsilon_{i}(f(x_i)-\pi_{n}f_0(x_i))|]\\ 
  &\lesssim \max_{1 \leq k \leq n} E[\sup_{f \in \mathcal{F}_n} |\frac{1}{\sqrt{k}}\sum_{i=1}^{k}\xi_{i}(f(x_i) - \pi_{n}f_0(x_i))|]\\
  & \lesssim \int_{0}^{\infty}H^{1/2}(u)du,
\end{align*}
where $H(u) = \log N(u,\mathcal{F}_{n}, ||\cdot||_{\infty})$, and $\xi_{1}, ...,\xi_{n}$ are i.i.d Rademacher random variable which are independent of $X_1,...,X_n.$
\end{lemma}

\begin{proof}
Since $||\epsilon||_{p,1} < \infty$ for some $p \geq 2$ and $E[\sup_{f\in \mathcal{F}_n}|f(X)|] < \infty$, the first inequality is a direct consequence of the multiplier inequalities(Lemma 2.9.1 in Van der Vaart and Wellnez(1996)).
For the second inequality, when $X_1, ..,X_n$ are given, $\lbrace \frac{1}{\sqrt{n}}\sum_{i=1}^{k}\xi_{i}(f(x_i) - \pi_{n}f_{0}(x_i)), f \in \mathcal{F}_n\rbrace$ is a Rademacher process and has a sub-guassian process. Then, it follows from Corollary 2.2.8 and the duality between packing and covering numbers.
\begin{align*}
    &E_{\xi | X}[\sup_{f \in \mathcal{F}_n}|\frac{1}{\sqrt{k}}\sum_{i=1}^{k}\epsilon_{i}(f(x_i)-\pi_{n}f_0(x_i))|]\\
    &\lesssim \int_{0}^{\infty}\sqrt{\log D(t,\mathcal{F}_n,||\cdot||_{k})} ||dt\\
    & \leq \int_{0}^{\infty}\sqrt{\log N(\frac{t}{2},\mathcal{F}_{n}, ||\cdot||_{k})}dt\\
    &\lesssim \int_{0}^{\infty}\sqrt{\log N(u,\mathcal{F}_{n}, ||\cdot||_{k})}du\\
    &= \int_{0}^{\infty}H^{1/2}(u)du.
\end{align*}
Note that the bound on the right hand side does not depend on $k$. Therefore, 
\begin{align*}
    &\max_{1 \leq k \leq n} E\left[\sup_{f\in \mathcal{F}_n}|\frac{1}{\sqrt{k}}\sum_{i=1}^{k}\xi_{i}(f(X_i)-\pi_{n}f_{0}(X_i))|\right]\\
    &=\max_{1 \leq k \leq n} E_{X}\left[E_{\xi|X}\left[\sup_{f\in \mathcal{F}_n}|\frac{1}{\sqrt{k}}\sum_{i=1}^{k}\xi_{i}(f(X_i)-\pi_{n}f_{0}(X_i))|\right]\right]\\
    &\lesssim \max_{1\leq k \leq n}E_{X}\left[ \int_{0}^{\infty} H^{1/2}(u) du\right]\\
    &= \int_{0}^{\infty}H^{1/2}(u)du.
\end{align*}
\end{proof}



In many applications, $\mcal{F}_n$ contains functions parameterized by some parameter $\theta\in K_n$, where $K_n$ is some compact set in $\mbb{R}^{p_n}$. Examples include linear sieves such as polynomials, splines and nonlinear sieves such as neural networks as will be discussed in Section \ref{Sec: NN}. We mainly focus on the $\ell_1$-penalty due to its ability in conducting feature selections. Specifically, for any $f\in\mcal{F}_n$, we consider
\begin{equation}\label{Eq: parameter penalty}
J_n(f)=\sum_{i=1}^{p_n}\abs{\theta_i}.
\end{equation}

\begin{mydef}\label{Def: well-defined}
 The map $J_n$ defined in (\ref{Eq: parameter penalty}) is said to be well-defined if $\theta_1\neq\theta_2$ produce the same function $f\in\mcal{F}_n$, then
$$
J_n(f(x,\theta_1))=J_n(f(x,\theta_2)).
$$
\end{mydef}

From now on, we assume that $J_n$ is well-defined. Next, we set
\begin{align*}
    \mbb{Q}_n(\theta) & =\mbb{Q}_n(f)=\frac{1}{n}\sum_{i=1}^n(Y_i-f(X_i,\theta))^2\\
    Q(\theta) & =Q(f)=E[(Y-f(X,\theta))^2].
\end{align*}
It is clear that $f_0$ minimizes $Q(f)$ and $Q(f)=Q(f_0)+\norm{f-f_0}^2$.

\begin{thm}\label{Thm: L1-penalty general result}
Let $\mcal{F}_n$ be a collection of real analytic functions and let $\Theta_n^*=\{\theta\in K_n:Q(\theta)=Q(\theta_n^*)\}$, where $\theta_n^*$ is a parameterization of $\pi_nf_0$. For any $\delta>0$,
$$
\lambda_n\left[J_n(\pi_nf_0)-J_n(\hat{f}_n)\right]\leq C_{\nu,\delta}\lambda_n\sqrt{p_n}\left(\frac{\log n}{\sqrt{n}}+\sqrt{p_n}\lambda_n^{\frac{\nu}{\nu-1}}+\norm{f_0-\pi_nf_0}^2+O(\eta_n)\right)^{1/\nu},
$$
with probability at least $1-\delta$.
\end{thm}

\begin{proof}
Since $\mcal{F}_n$ contains real analytic functions, by the Lojasiewicz inequality \cite{ji1992global}, there exist $C>0$ and $\nu>0$ such that
$$
\inf_{\tilde{\theta}\in\Theta_n^*}\norm{\theta-\tilde{\theta}}\leq C\abs{Q(\theta)-Q(\theta^*)}^{1/\nu},\quad\forall\theta\in K_n.
$$
Let $\hat{\theta}_n$ be a parameterization of $\hat{f}_n$ and define $\vartheta_n=\mrm{argmin}_{\theta\in\Theta_n^*}\norm{\theta-\hat{\theta}_n}$. Then note that $\inf_{\theta\in\Theta_n^*}\norm{\theta-\hat{\theta}_n}=\norm{\vartheta_n-\hat{\theta}_n}$, we obtain
\begin{align*}
    \norm{\vartheta_n-\hat{\theta}_n}^\nu & \lesssim\abs{Q(\hat{\theta}_n)-Q(\vartheta_n)}=\abs{Q(\hat{\theta}_n)-Q(\theta_n^*)}\\
    & =\abs{Q(\hat{\theta}_n)-Q(f_0)-\norm{f_0-\pi_nf_0}^2}\\
    & \lesssim Q(\hat{\theta}_n)-Q(f_0)+\norm{f_0-\pi_nf_0}^2\\
    & \lesssim Q(\hat{\theta}_n)-Q(\theta_n^*)+\norm{f_0-\pi_nf_0}^2\\
    & =Q(\hat{\theta}_n)-Q(\vartheta_n)+\norm{f_0-\pi_nf_0}^2\\
    & =Q(\hat{\theta}_n)-\mbb{Q}(\hat{\theta}_n)-(Q(\vartheta_n)-\mbb{Q}(\vartheta_n))+\mbb{Q}_n(\hat{\theta}_n)-\mbb{Q}_n(\vartheta_n)+\norm{f_0-\pi_nf_0}^2
\end{align*}
By Lemma 3.3 in \cite{ji1992global}, for any $\delta>0$,
$$
\abs{\mbb{Q}_n(\theta)-Q(\theta)}\lesssim_\delta\frac{\log n}{\sqrt{n}},\quad\forall\theta\in K_n.
$$
with probability at least $1-\delta$. On the other hand, by the definition of $\hat{f}_n$, 
\begin{align*}
\mbb{Q}_n(\hat{\theta}_n)-\mbb{Q}_n(\pi_nf_0) & \leq\lambda_n\left[J_n(\pi_nf_0)-J_n(\hat{f}_n)\right]+O(\eta_n)\\ 
    & \leq\lambda_n\sum_{i=1}^{p_n}\abs{\vartheta_{n,i}-\hat{\theta}_{n,i}}\\
    & \leq\lambda_n\sqrt{p_n}\norm{\vartheta_n-\hat{\theta}_n}\\
    & \leq\frac{1}{2}\norm{\vartheta_n-\hat{\theta}_n}^\nu+\frac{2(\nu-1)\sqrt{p_n}}{\nu^{1+1/(\nu-1)}}\lambda_n^{\nu/(\nu-1)},
\end{align*}
where the last inequality follows from the Young's inequality. Therefore, we have with probability at least $1-\delta$,
\begin{align*}
    \norm{\vartheta_n-\hat{\theta}_n}^\nu & \lesssim_{\nu,\delta}\frac{\log n}{\sqrt{n}}+\frac{1}{2}\norm{\vartheta_n-\hat{\theta}_n}^\nu+\sqrt{p_n}\lambda_n^{\frac{\nu}{\nu-1}}+\norm{f_0-\pi_nf_0}^2+O(\eta_n),
\end{align*}
which implies that
\begin{align*}
    \lambda_n\left[J_n(\pi_nf_0)-J_n(\hat{f}_n)\right] & \leq\lambda_n\sqrt{p_n}\norm{\vartheta_n-\hat{\theta}_n}\\
    & \lesssim_{\nu,\delta}\lambda_n\sqrt{p_n}\left(\frac{\log n}{\sqrt{n}}+\sqrt{p_n}\lambda_n^{\frac{\nu}{\nu-1}}+\norm{f_0-\pi_nf_0}^2+O(\eta_n)\right)^{1/\nu},
\end{align*}
with probability at least $1-\delta$.
\end{proof}

As a direct consequence of Lemma \ref{Lm: Basic Ineq}, Lemma \ref{Lm: Convergence of Multiplier Process} and Theorem \ref{Thm: L1-penalty general result}, the following general result on consistency of penalized least square with $J_n(f)$ being defined in (\ref{Eq: parameter penalty}) can be obtained.

\begin{thm}\label{Thm: consistency PLS general result}
Suppose that $\mcal{F}_n$ is a collection of real analytic function such that $\mcal{F}_n\subset\mcal{F}_{n+1}$ and $\mcal{F}=\bigcup_{n=1}^\infty\mcal{F}_n$. Let $\Theta_n^*=\{\theta\in K_n:Q(\theta)=Q(\theta_n^*)\}$, where $\theta_n^*$ is a parameterization of $\pi_nf_0$. Under the following conditions
\begin{equation}\label{Eq: Entropy Condition}
\int_0^\infty H^{1/2}(u) du=o(\sqrt{n}),\quad\mrm{as }n\to\infty.
\end{equation}
Then for any $\delta>0$, there exists $\nu>0$ such that
$$
\norm{\hat{f}_n-f_0}_n\to0,\quad\mrm{as }n\to\infty,
$$
with probability at least $1-\delta$ provided that $\lambda_n=o\left(p_n^{-1/2}\wedge p_n^{-(\nu-1)/(2\nu)}\right)$.
\end{thm}

\section{Neural networks}\label{Sec: NN}
In this section, we will apply general consistency theorem into neural networks models. we mainly consider two activation function: Tanh and ReLu.

We now consider neural network regression function estimators. Given the
training sequence $D_n = \{(X_1, Y_1), . . . , (X_n, Y_n)\}$ of $n$ i.i.d copies of $(X, Y )$
the parameters of the network are chosen to minimize the empirical $L_2$ risk with penalty terms
\begin{equation}
    \frac{1}{n}\sum_{i=1}^{n}(f(X_i)-Y_i)^{2} + \lambda_{n}J_{n}(f)
\end{equation}
In order to avoid unidentifiability issue, We restrict the choice of parameters. We focus on the sieve of neural networks with one hidden layer. 
\begin{equation}\label{Eq: NN Sieve}
\begin{split}
\mathcal{F}_{r_n} &= \lbrace \alpha_0 + \sum_{j = 1}^{r_n} \alpha_j \sigma \left( \gamma_{j}^{T} x + \gamma_{0,j} \right) : \mbf{\gamma}_{j} \in \mathbb{R}^d, \alpha_j, \gamma_{0,j} \in \mathbb{R}, \sum_{j = 0}^{r_n} \vert \alpha_j\vert \leq V_n \\
& \text{ for some } V_n > 1 \text{ and } \max_{1 \leq j \leq r_n} \sum_{i=0}^d \vert \gamma_{i,j} \vert \leq M_n  \text{ for some } M_n > 0 \rbrace,
\end{split}
\end{equation}
where 
$
r_n, V_n, M_n \rightarrow \infty \text{ as } n \rightarrow \infty.
$

\subsection{Neural networks with tanh function as activation function}
%

As have been pointed out in \cite{fukumizu1996regularity} and \cite{Fukumizu2003LikelihoodNetworks}, when the number of hidden units in neural networks are unknown, the parameters are unidentifiable. On the other hand, based on the following result due to \cite{sussmann1992uniqueness}, such unidentifiability mainly comes from either the signed permuations in weights and biases or from the model with higher dimensional parameter space containing zero weights or biases. 

\begin{mydef}
 A neural network is said to be minimal if no networks with fewer hidden units have the same input-output map.
\end{mydef}

\begin{thm}[Sussman (1992)]\label{Thm: minimal nn}
A neural network with $r$ hidden units is minimal if and only if 
\begin{enumerate}
    \item $\gamma_j\neq 0$, for all $j=1,\ldots, r$.
    \item $[\alpha_1,\ldots,\alpha_r]^T\neq0$.
    \item For any two different indices $j_1$ and $j_2$, $(\gamma_{j_1}^T,\gamma_{0,j_1})\neq\pm(\gamma_{j_2}^T,\gamma_{0,j_2})$.
\end{enumerate}
\end{thm}

\begin{prop}
Let $\mcal{F}_{r_n}$ be as defined in (\ref{Eq: NN Sieve}) with $\sigma(\cdot)=\tanh(\cdot)$. Then the map
\begin{align*}
    J_n: \mcal{F}_{r_n} & \to\mbb{R}\\
          f & \mapsto\sum_{j=0}^{r_n}\abs{\alpha_j}+\sum_{i=0}^d\sum_{j=1}^{r_n}\abs{\gamma_{i,j}},
\end{align*}
is well-defined.
\end{prop}

\begin{proof}
For any $f\in\mcal{F}_{r_n}$, let $\theta_{n0}=[\alpha_0,\ldots,\alpha_{r_n'}, \gamma_1^T,\ldots,\gamma_{r_n'}^T,\gamma_{0,1},\ldots,\gamma_{0,r_n'}]^T$ where $r_n'\leq r_n$ be the minimal network parameterization. We now consider two cases:
\begin{itemize}
    \item[] Case 1. $r_n'=r_n$.\\
            In this case, by Theorem \ref{Thm: minimal nn}, different parameterizations produce the same function $f$ only when there exists two different indices $j_1$ and $j_2$ such that $(\gamma_{j_1}^T,\gamma_{0,j_1})=\pm(\gamma_{j_2}^T,\gamma_{0,j_2})$. Note that such difference does not change the value of $\sum_{i=0}^d\sum_{j=0}^{r_n}\abs{\gamma_{i,j}}$ since $\tanh$ is an odd function and hence $J_n$ is well-defined in this case.
    \item[] Case 2. $r_n'<r_n$.\\
            In this case
            $$
            J_n(f(x,\theta_{n0}))=\sum_{i=0}^{r_n'}\abs{\alpha_j}+\sum_{i=0}^d\sum_{j=1}^{r_n'}\abs{\gamma_{i,j}}.
            $$
            Suppose that $\theta_n\in\mbb{R}^{r_n(d+2)+1}$ is a parameterization that produces the same $f$, by Theorem \ref{Thm: minimal nn}, there are three possible scenarios.
            \begin{enumerate}
                \item There exist $\{j_1,\ldots,j_{r_n-r_n'}\}\subset\{1,\ldots,r_n\}$ such that $\gamma_{j_1}=\gamma_{j_2}=\cdots=\gamma_{j_{r_n-r_n'}}=\mbf{0}$. Through suitable rearrangement of the indices, without loss of generality, assume that $j_1=r_n'+1, \ldots, j_{r_n-r_n'}=r_n$. On the other hand, since
                \begin{align*}
                    \sum_{i=0}^d\sum_{j=1}^{r_n}\abs{\gamma_{i,j}} & =\sum_{i=0}^d\sum_{j=1}^{r_n'}\abs{\gamma_{i,j}}+\sum_{i=0}^d\sum_{j=1+r_n'}^{r_n}\abs{\gamma_{i,j}}=\sum_{i=0}^d\sum_{j=1}^{r_n'}\abs{\gamma_{i,j}},
                \end{align*}
                such different parametrization does not change the value of the second term in the definition of $J_n$.
                
                \item There exist $\{j_1,\ldots,j_{r_n-r_n'}\}\subset\{1,\ldots,r_n\}$ such that $[\alpha_{j_1},\ldots,\alpha_{j_{r_n-r_n'}}]^T=0$. Similarly, without loss of generality, we may assume that $j_1=r_n'+1, \ldots, j_{r_n-r_n'}=r_n$. But note that
                \begin{align*}
                    \sum_{j=0}^{r_n}\abs{\alpha_j} & =\sum_{j=0}^{r_n'}\abs{\alpha_j}+\sum_{j=1+r_n'}^{r_n}\abs{\alpha_j}=\sum_{j=0}^{r_n'}\abs{\alpha_j},
                \end{align*}
                such different parametrization does not change the value of the first term in the definition of $J_n$.
                
                \item There exist two indices $j_1$ and $j_2$ such that $(\mbf{\gamma}_{j_1}^T,\gamma_{0,j_1})=\pm(\mbf{\gamma}_{j_2},\gamma_{0.j_2})$. The reasoning for such scenario is exactly the same as those discussed in Case 1.
            \end{enumerate}
\end{itemize}
Therefore, it can be concluded that $J_n$ is indeed well-defined. 
\end{proof}

Now, we are ready to state and prove the consistency theorem for neural network sieve estimators based on Theorem \ref{Thm: consistency PLS general result}.

\begin{thm}
    Let $\mcal{F}_{r_n}$ be as defined in (\ref{Eq: NN Sieve}) with $\sigma(\cdot)=\tanh(\cdot)$. Under the assumptions
    \begin{equation}\label{Eq: NN-condition}
    r_nV_n^2\log(r_nV_n)=o(n),\quad \mrm{as }n\to\infty,
    \end{equation}
    and $\lambda_n=o(r_n^{-1/2}\wedge r_n^{-(\nu-1)/(2\nu)})$, then for any $\delta>0$,
    $$
    \norm{\hat{f}_n-f_0}_n\to0,\quad \mrm{as }n\to\infty,
    $$
    with probability at least $1-\delta$.
\end{thm}

\begin{proof}
    According to Theorem 14.5 in \cite{anthony1999neural},
    $$
    N(\epsilon,\mcal{F}_{r_n},\norm{\cdot}_\infty)\leq\left(\frac{2e[r_n(d+2)+1]V_n^2}{\epsilon(V_n-1)}\right)^{r_n(d+2)+1}:=A_{r_n,V_n,d}\epsilon^{-[r_n(d+2)+1]},
    $$
    where $A_{r_n,V_n,d}=(2e[r_n(d+2)+1]V_n^2/(V_n-1))^{r_n(d+2)+1}$. Let
    \begin{align*}
        B_{r_n,V_n,d} & =\log A_{r_n,V_n,d}-[r_n(d+2)+1]=[r_n(d+2)+1]\left(\log\frac{2e[r_n(d+2)+1]V_n^2}{V_n-1}-1\right)\\
        & =[r_n(d+2)+1]\log\frac{2[r_n(d+2)+1]V_n^2}{V_n-1}.
    \end{align*}
    Then
    \begin{align*}
        H(u) & =\log N(u,\mcal{F}_{r_n},\norm{\cdot}_\infty)\\
            & =\log A_{r_n,V_n,d}+[r_n(d+2)+1]\log\frac{1}{u}\\
            & \leq\log A_{r_n,V_n,d}-[r_n(d+2)+1]+]+[r_n(d+2)+1]\frac{1}{u}\\
            & =B_{r_n,V_n,d}\left(1+\frac{1}{u}\right),
    \end{align*}
    where the last inequality follows by noting that $V_n^2-V_n+1\geq0$ for all $V_n$ so that $\log\frac{2[r_n(d+2)+1]V_n^2}{V_n-1}\geq\log\frac{8(V_n-1)}{V_n-1}=\log8\geq1$. Moreover,
    \begin{align*}
        \int_0^\infty H^{1/2}(u)du & \leq B_{r_n,V_n,d}^{1/2}\int_0^{2V_n}\left(1+\frac{1}{u}\right)^{1/2}du\\
        & =B_{r_n,V_n,d}^{1/2}\left[\int_0^1\left(1+\frac{1}{u}\right)^{1/2}du+\int_1^{2V_n}\left(1+\frac{1}{u}\right)^{1/2}du\right]\\
        & \leq B_{r_n,V_n,d}^{1/2}\left[\sqrt{2}\int_0^1u^{-1/2}du+\sqrt{2}(2V_n-1)\right]\\
        & \leq4\sqrt{2}B_{r_n,V_n,d}^{1/2}V_n\\
        & \asymp_d \left(r_nV_n^2\log(r_nV_n)\right)^{1/2}.
    \end{align*}
    Therefore, the entropy condition (\ref{Eq: Entropy Condition}) in Theorem \ref{Thm: consistency PLS general result} is satisfied based on (\ref{Eq: NN-condition}). The result then follows from Theorem \ref{Thm: consistency PLS general result} by noting that the total number of parameters in a neural network with $r_n$ hidden units is $r_n(d+2)+1$.
\end{proof}

\subsection{Neural networks with ReLU function as activation function}
Rectified Linear Unit (ReLU) is one of commonly used activation function in neural networks. However, the theory on minimal representation of shallow ReLU network, which was developed in \cite{dereich2022minimal}, is much more complicated than its counterpart for tanh activation function. Therefore, it is difficult for the map $J_n:f\in\mcal{F}_{r_n}\mapsto\sum_{j=0}^{r_n}\abs{\alpha_j}+\sum_{i=0}^d\sum_{j=1}^{r_n}\abs{\gamma_{i,j}}$ to be well-defined. On the other hand, the $\ell_1$ regularization is well-known to produce sparse model in statistics \cite{tibshirani1996regression, hastie2015statistical}. Therefore, one way to address this issue is to consider its nonparametric counterpart. To be specific, we consider the following penalty function as mentioned in \cite{rosasco2013nonparametric}:
\begin{align*}
    J_n: \mcal{F}_{r_n} & \to \mbb{R}^+\\
    f & \mapsto \frac{1}{n}\sum_{i=1}^n\norm{\nabla f(X_i)}_1.\numberthis\label{Eq: nonparametric sparsity penalty}
\end{align*}
Moreover, due to the positive homogeneity property of ReLU function:
$$
\sigma(cx)=c\sigma(x)\mrm{ for any }c\geq0,
$$
we may assume, without loss of generality, that the ReLU network in $\mcal{F}_{r_n}$ has the form
\begin{equation}\label{Eq: RELU NN Sieve}
\sum_{j=1}^{r_n}\alpha_j\sigma\left(\gamma_j^Tx+\gamma_{0,j}\right),
\end{equation}
where $\alpha_j\in\{-1,1\}$ and $\gamma_j,\gamma_{0,j}$ have the same restrictions as in (\ref{Eq: NN Sieve}).

Note that for any $f=\alpha_0+\sum_{j=1}^{r_n}\alpha_j\sigma\left(\gamma_j^Tx+\gamma_{0,j}\right)\in\mcal{F}_{r_n}$, the partial derivative of $f$ with respect to $x^{(k)}$, the $k$th component in $x$, is given by
$$
\frac{\partial f}{\partial x^{(k)}}=\sum_{j=1}^{r_n}\alpha_j\gamma_{j}^{(k)}\mbb{I}_{\left\{\gamma_j^Tx+\gamma_{0,j}>0\right\}},
$$
which implies that
\begin{align*}
    \norm{\nabla f(x)}_1 & =\sum_{k=1}^d\abs{\frac{\partial f}{\partial x^{(k)}}}\leq\sum_{k=1}^d\sum_{j=1}^{r_n}\abs{\alpha_j}\abs{\gamma_j^{(k)}}\\
    & =\sum_{k=1}^d\sum_{j=1}^{r_n}\abs{\gamma_j^{(k)}}.
\end{align*}
Therefore, the regular $\ell_1$-penalty dominates the penalty term defined in (\ref{Eq: nonparametric sparsity penalty}) and hence a minimizer for the optimization problem with the penalty in (\ref{Eq: nonparametric sparsity penalty}) is also a minimizer for the optimization problem with the $\ell_1$-penalty.

\begin{thm}
    Let $\mcal{F}_{r_n}$ be as defined in (\ref{Eq: RELU NN Sieve}) with $\sigma(\cdot)=ReLU(\cdot)$. Under the assumptions
    \begin{equation}\label{Eq: ReLU entropy condition}
        r_n^3M_n^2\log r_n=o(n),\quad \mrm{as }n\to\infty,
    \end{equation}
    and $\lambda_n=o((r_nM_n)^{-1})$, then
    $$
    \norm{\hat{f}_n-f_0}_n\xrightarrow{p}0,\quad \mrm{as }n\to\infty.
    $$
\end{thm}

\begin{proof}
    Based on Lemma 5 in \cite{schmidt2020nonparametric}, we have
    $$
    H(u)\leq [r_n(d+2)+2]\log\frac{16(d+1)^2(r_n+1)^2}{u}.
    $$
    Moreover, since $\log x\leq x-1$ for $x>0$, we have
    \begin{align*}
        H(u) & \leq [r_n(d+2)+2]\left(2\log(4(d+1)(r_n+1))+\log\frac{1}{u}\right)\\
        & \leq [r_n(d+2)+2]\left(2\log(4(d+1)(r_n+1))+\frac{1}{u}-1\right)\\
        & \leq 2[r_n(d+2)+2]\log\frac{4(d+1)(r_n+1)}{e^{1/2}}\left(1+\frac{1}{u}\right)\\
        & :=C_{r_n,d}\left(1+\frac{1}{u}\right).
    \end{align*}
    Therefore,
    \begin{align*}
        \int_0^\infty H^{1/2}(u)du & \leq C_{r_n,d}^{1/2}\int_0^{2r_nM_n}\left(1+\frac{1}{u}\right)^{1/2}du\\
        & =C_{r_n,d}^{1/2}\left[\int_0^1\left(1+\frac{1}{u}\right)^{1/2}du+\int_1^{2r_nM_n}\left(1+\frac{1}{u}\right)^{1/2}du\right]\\
        & \leq C_{r_n,d}^{1/2}\left[\sqrt{2}\int_0^1u^{1/2}du+\sqrt{2}(2r_nM_n-1)\right]\\
        & \leq 4\sqrt{2}C_{r_n,d}^{1/2}r_nM_n\\
        & \asymp_d \left(r_n^3M_n^2\log r_n\right)^{1/2}
    \end{align*}
    Therefore, the entropy condition (\ref{Eq: Entropy Condition}) in Theorem 2 is satisfied based on (\ref{Eq: ReLU entropy condition}). 
    
    Next, note that
    \begin{align*}
        \sup_{f_1,f_2\in\mcal{F}_{r_n}}\abs{J_n(f_1)-J_n(f_2)} & =\sup_{f_1,f_2\in\mcal{F}_{r_n}}\abs{\frac{1}{n}\sum_{i=1}^n\sum_{k=1}^d\left(\abs{\frac{\partial f_1}{\partial x^{(k)}}(X_i)}-\abs{\frac{\partial f_2}{\partial x^{(k)}}(X_i)}\right)}\\
        & \leq 2r_nM_n,
    \end{align*}
    we can know that if $\lambda_n=o((r_nM_n)^{-1})$, then $\lambda_n[J_n(\pi_nf_0)-J_n(\hat{f}_n)]\to0$ as $n\to\infty$ and the desired result follows from the basic inequality in Lemma \ref{Lm: Basic Ineq}.
\end{proof}

\section{Simulation}
To validate the consistency of penalized neural networks, we run simulations with three different nonlinear functions.
The response was simulated through the following equation:
\begin{equation}
    y_i = f_0(x_i) + \epsilon_i, i = 1,.., n,
\end{equation}
where $x_1,.., x_n \sim Uniform(-2,2), \epsilon_1, ..., \epsilon_n \stackrel{i.i.d}{\sim} \mathcal{N}(0, 0.7^2)$. For the true function $f_0$, we consider three different nonlinear functions:
\begin{enumerate}
    \item a neural network with one single hidden layer and two hidden units,
    \item A trigonometric function: $$f_0 = sin(\frac{\pi x}{3}) + \frac{1}{3} cos(\frac{\pi x }{4}+ 1),$$
    \item a complex nonlinear function: $$f_0 = sin(x)+exp(-4x^{2}).$$
\end{enumerate}
We then trained a neural network using the gradient descent algorithm and set the number of iterations as 20,000. To accommodate the assumption of tuning parameter $\lambda_n$, we took $\lambda_n = \frac{\lambda}{n}$, where $\lambda= 10$. We chose 5, 10, 15, 20 number of hidden nodes to train penalized neural networks.  five different sample sizes: 100, 200, 500, 1000 and 2000 were chosen. We compared
the errors $\|\hat{f}_n-f_0\|_n^2= \frac{1}{n} \sum_{i=1}^{n}(\hat{f}(x_i) - f_0(x_i))^2$
 and the least square errors $\mbb{Q}_n(\hat{f}_n)= \frac{1}{n}\sum_{i=1}^{n}(\hat{f}(x_i)-y_i)^{2}$ under four different
sample sizes. The code can be found on 
\url{https://github.com/linjack121/consistency-of-penalized-neural-network}. This is computed by local desktop with Intel Core i7 CPU.
\subsection{Activation function: tanh} 
\begin{figure}
    \centering
    \includegraphics[scale = 0.35]{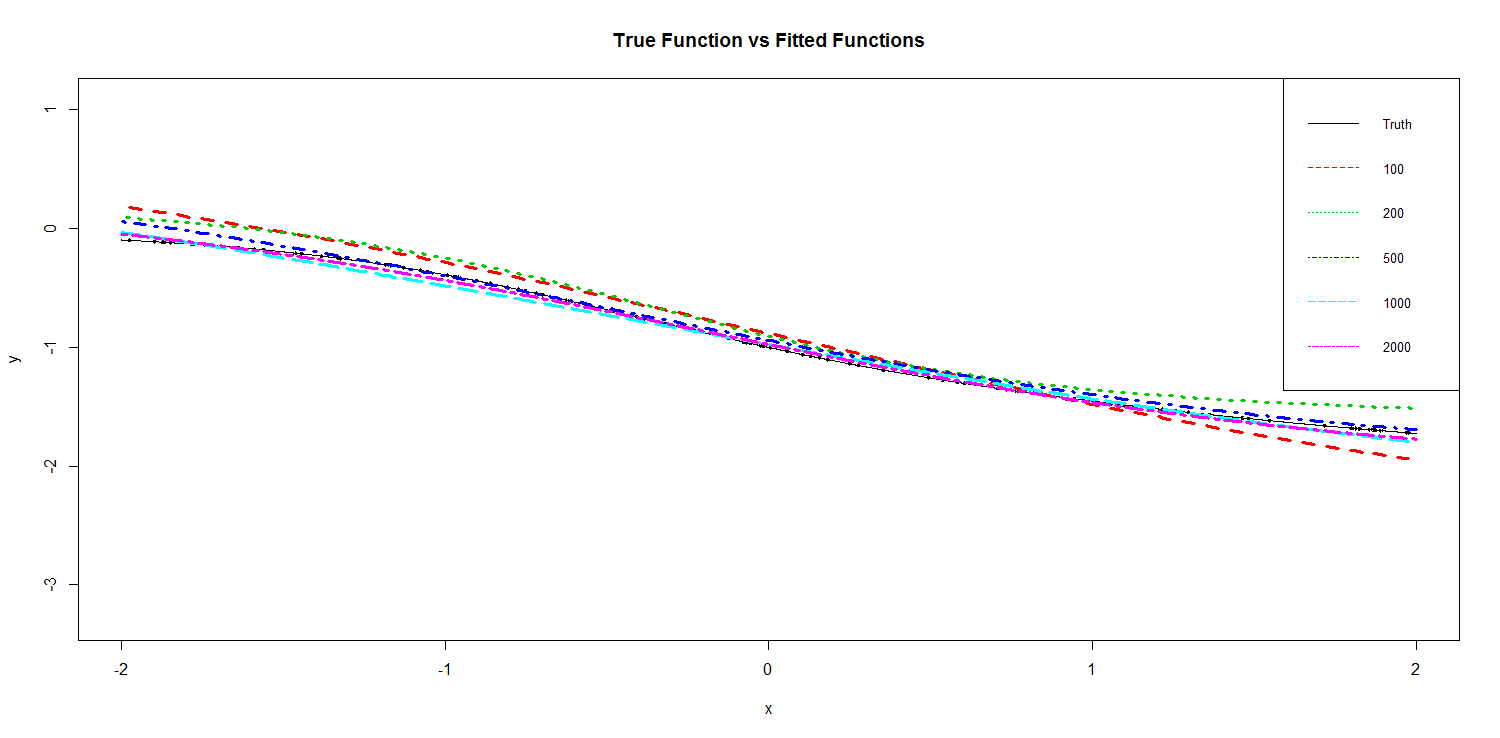}
    \caption{Tanh: Comparison between the true function $f_0$ and fitted functions under different sample sizes, where $f_0$ is a neural network with one single hidden layer and two hidden units.}
    \label{fig:my_label1}
\end{figure}

\begin{figure}
    \centering
    \includegraphics[scale = 0.35]{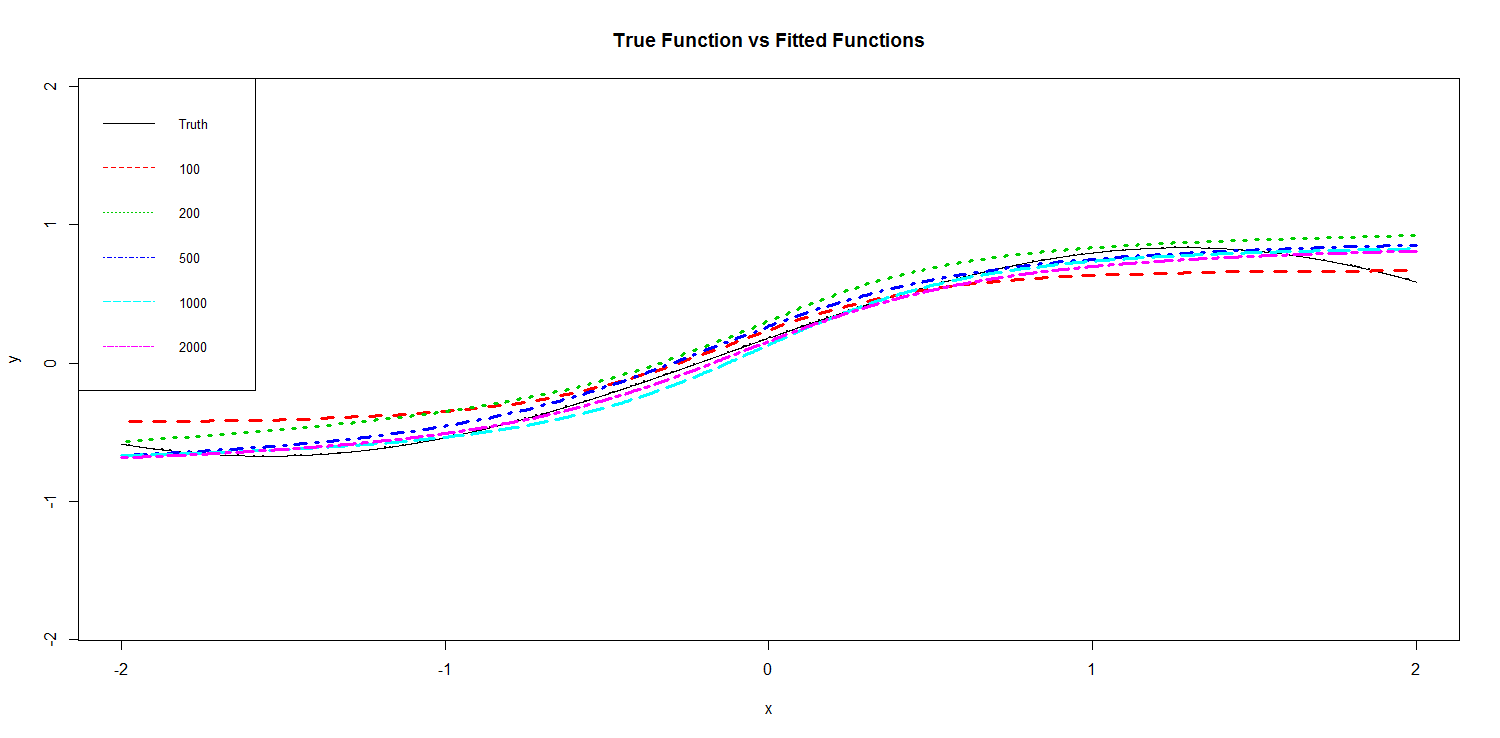}
    \caption{Tanh: Comparison between the true function $f_0 = sin(\frac{\pi x}{3}) + \frac{1}{3} cos(\frac{\pi x }{4}+ 1)$ and fitted functions under different sample sizes.}
    \label{fig:my_label2}
\end{figure}

\begin{figure}
    \centering
    \includegraphics[scale = 0.35]{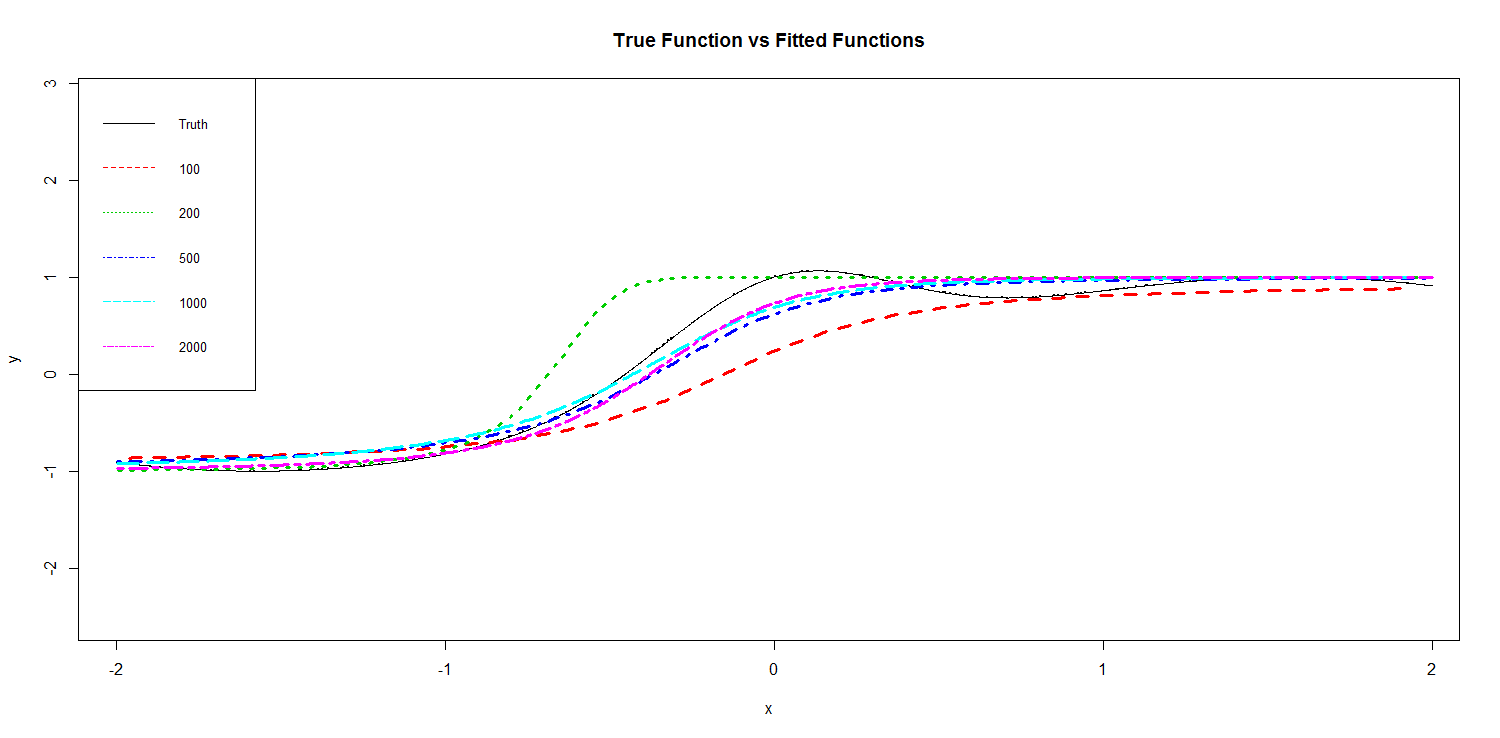}
    \caption{Tanh: Comparison between the true function $f_0 = sin(x)+exp(-4x^2)$ and fitted functions under different sample sizes.}
    \label{fig:my_label3}
\end{figure}
In this section, we simulate neural networks with tanh as activation function.  Three nonlinear functions are considered. From Figure 1 to Figure 3, the fitted curve is closer to the true function as the sample increases.

\begin{table}[h!]
\caption{Tanh: Comparison of errors $\|\hat{f}_n-f_0\|_n^2$ and the least square errors $\mbb{Q}_n(\hat{f}_n)$ after 20,000 iterations under different sample sizes.}\label{Tab: ConsistencyErr1NN}
\begin{center}
\begin{tabular}{lccccccccc}
\hline
\multirow{2}{*}{Sample Sizes} &  \multicolumn{2}{c}{Neural Network} & & \multicolumn{2}{c}{trigonometric function} & & \multicolumn{2}{c}{a complex function}\\
\cline{2-3}\cline{5-6}\cline{8-9}
	&  $\|\hat{f}_n-f_0\|_n^2$ & $\mbb{Q}_n(\hat{f}_n)$ & & $\|\hat{f}_n-f_0\|_n^2$ & $\mbb{Q}_n(\hat{f}_n)$ & & $\|\hat{f}_n-f_0\|_n^2$ & $\mbb{Q}_n(\hat{f}_n)$\\
 \hline
 100  &1.96E-2  &0.4394   & &2.21E-2  &0.439  & &9.31E-2   &0.607\\
 200  &1.80E-2  &0.4919   & &2.05E-2   &0.491   & &6.82E-2   &0.515  \\
 500  &3.20E-3  &0.4771   & &5.76E-3   &0.477  & &2.39E-2   &0.568 \\
 1000 &2.32E-3  &0.469  & &4.08E-3  &0.469  & &1.82E-2   &0.484 \\
 2000 &8.86E-4 &0.500  & &4.01E-3  &0.501  & &1.51E-2   &0.499\\
 \hline
\end{tabular}
\end{center}
\end{table}
Based on the result of Table 1, the errors $\|\hat{f}_n-f_0\|_n^2$  have a decreasing
pattern as the sample size increases. $\mbb{Q}_n(\hat{f}_n)$ oscillate around $0.49$ under 5 sample sizes as expected.  Figure 1-3 visualize the fitted functions and the true function which validate the consistency of penalized neural networks. 

\subsection{Activation function: ReLu} 
In this section, we consider penalized neural networks with rectified linear units(ReLU) as activation function. Same three nonlinear functions are used. The results are similar. Figure 4 to Figure 6 shows that  the estimate gets closer and closer to the true value of the parameter as the sample size increases from 100 to 2000. To quantify how closer the fitted curve to true function, Table 2 provides the errors $\|\hat{f}_n-f_0\|_n^2$ under 5 different sample sizes. For three different nonlinear functions, $\mbb{Q}_n(\hat{f}_n)$ also oscillate around $0.49$ under 5 sample sizes as expected
\begin{figure}
    \centering
    \includegraphics[scale = 0.35]{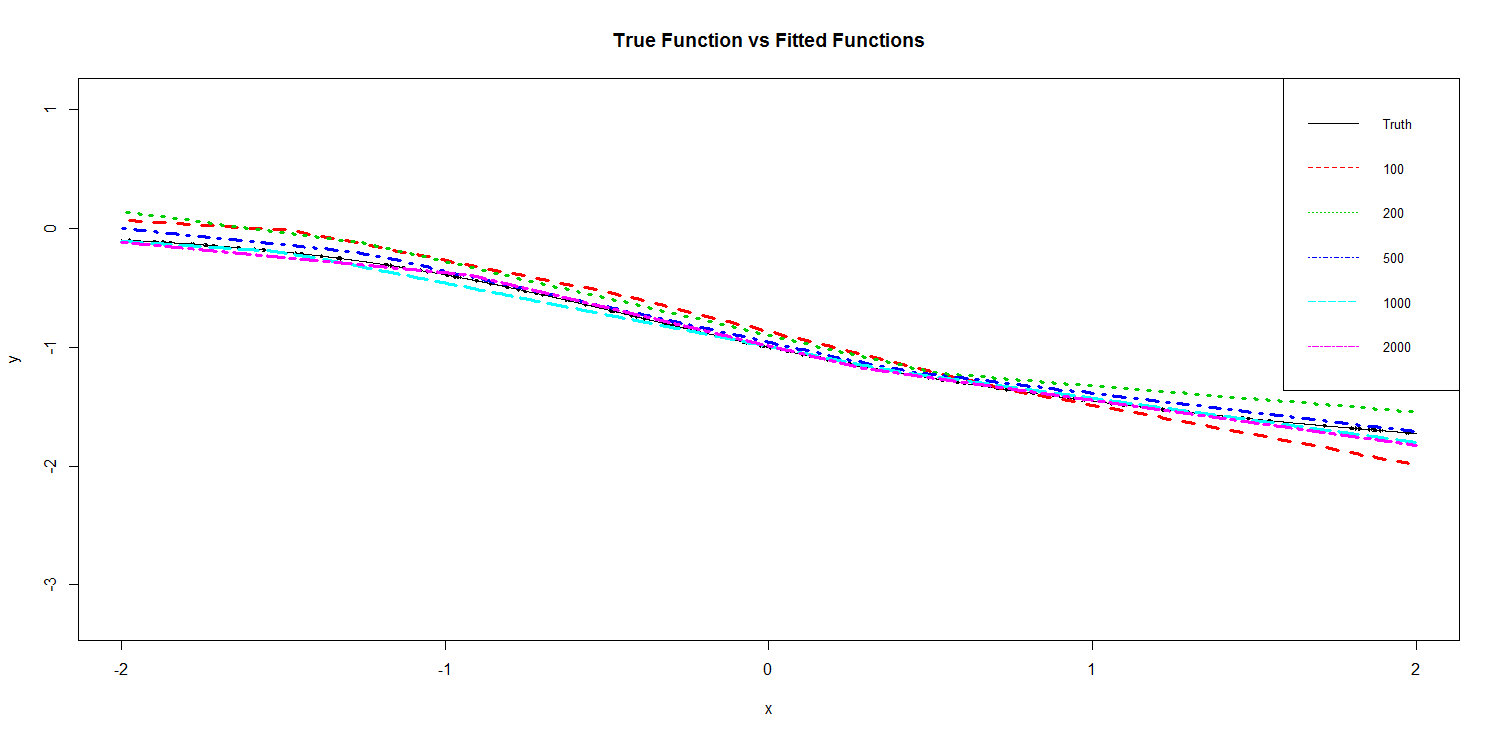}
    \caption{ReLu: Comparison between the true function $f_0$ and fitted functions under different sample sizes, where $f_0$ is a neural network with one single hidden layer and two hidden units.}
    \label{fig:my_label4}
\end{figure}

\begin{figure}
    \centering
    \includegraphics[scale = 0.35]{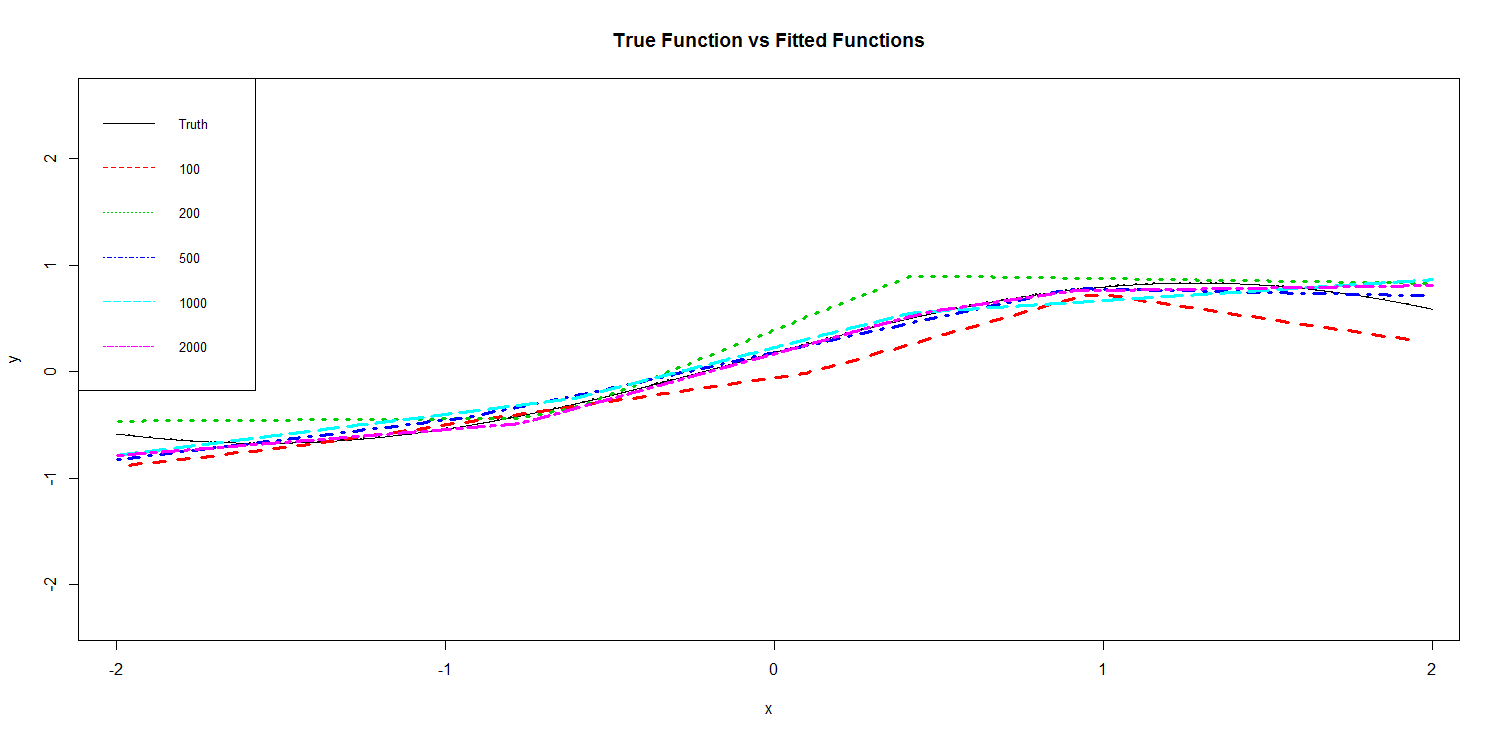}
    \caption{ReLu: Comparison between the true function $f_0 = sin(\frac{\pi x}{3}) + \frac{1}{3} cos(\frac{\pi x }{4}+ 1)$ and fitted functions under different sample sizes.}
    \label{fig:my_label5}
\end{figure}

\begin{figure}
    \centering
    \includegraphics[scale = 0.35]{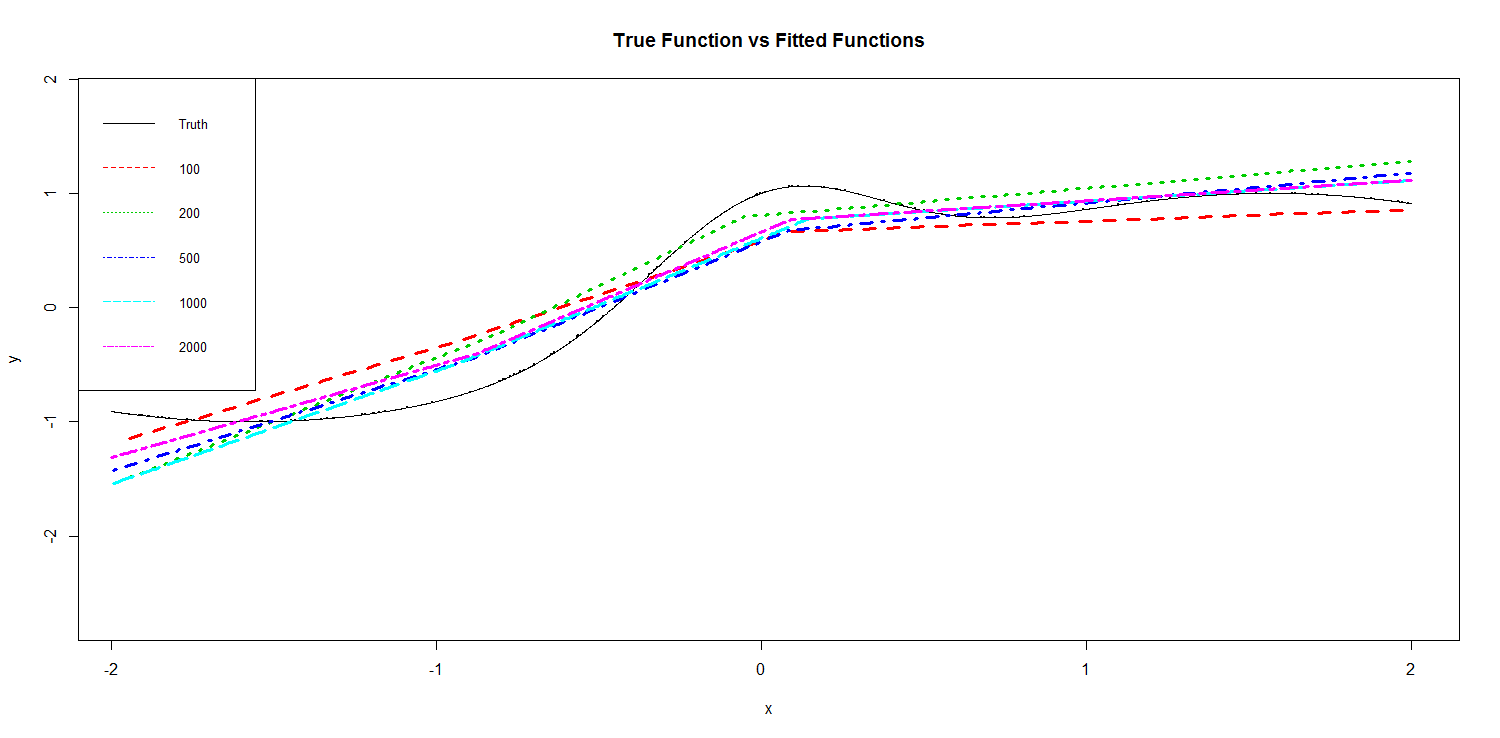}
    \caption{Comparison between the true function $f_0 = sin(x)+exp(-4x^2)$ and fitted functions under different sample sizes.}
    \label{fig:my_label6}
\end{figure}

\begin{table}[h!]
\caption{ReLu: Comparison of errors $\|\hat{f}_n-f_0\|_n^2$ and the least square errors $\mbb{Q}_n(\hat{f}_n)$ after 20,000 iterations under different sample sizes.}
\begin{center}
\begin{tabular}{lccccccccc}
\hline
\multirow{2}{*}{Sample Sizes} &  \multicolumn{2}{c}{Neural Network} & & \multicolumn{2}{c}{trigonometric function} & & \multicolumn{2}{c}{a complex function}\\
\cline{2-3}\cline{5-6}\cline{8-9}
	&  $\|\hat{f}_n-f_0\|_n^2$ & $\mbb{Q}_n(\hat{f}_n)$ & & $\|\hat{f}_n-f_0\|_n^2$ & $\mbb{Q}_n(\hat{f}_n)$ & & $\|\hat{f}_n-f_0\|_n^2$ & $\mbb{Q}_n(\hat{f}_n)$\\
 \hline
 100  &2.04E-2  &0.439   & &3.39E-2   &0.607  & &7.47E-2   & 0.451\\
 200  &1.86E-2  &0.491   & &3.42E-2   &0.515  & &6.35E-2   &0.480  \\
 500  &2.56E-3  &0.477   & &4.86E-3   &0.569  & &5.00E-2  &0.560 \\
 1000 &1.15E-3  &0.469   & &9.44E-3   &0.484  & &4.90E-2   &0.465 \\
 2000 &1.07E-3  &0.500   & &3.66E-3   &0.499  & &3.76E-2   &0.499\\
 \hline
\end{tabular}
\end{center}
\end{table}

\section{Conclusion}
With the success of deep learning in various areas, neural networks have regained their popularity in making accurate predictions. On the other hand, due to the difficulty of interpretations, neural networks are often regarded as ``black boxes". Recently, many researchers have started to focus on the interpretation of neural networks. For example, a goodness-of-fit test has been proposed by \cite{shen2021goodness}, which can be used to determine whether adding an additional input variable to the model is beneficial or not. In \cite{horel2020significance}, a significance test was proposed to test whether an input variable is statistically significant or not. 

The aforementioned research both considered the setting of classical nonparametric least square. However, in practice, neural networks are often trained based on certain regularization techniques. Therefore, it is worthwhile to understand the statistical properties for regularized neural networks, which is the motivation of this paper. In this paper, we mainly consider the consistency of neural networks with regularization, which is fundamental for further investigation on statistical properties such as rate of convergence as well as significance test based on regularized neural network estimates. 

A general result about consistency on nonparametric penalized least square setting is derived in this paper, which can be adapted to develop consistency for various deep neural networks. On the other hand, most regularization in practice are based on $\ell_1$ or $\ell_2$-penalty and neural networks, even with a single hidden layer, are well-known for their unidentifiability in their parameters. Theories on networks with tanh activation functions are easier to derive due to the simple conditions for minimal tanh networks. But this is not the case for other commonly used activation functions. We have demonstrated one possible way by considering a nonparametric version of sparse penalty as a counterpart for the commonly used $\ell_1$-regularization.


\label{sec:conc}
\newpage
\typeout{}
\bibliographystyle{plain}
\bibliography{main.bib}

\end{document}